\documentclass[twocolumn,letterpaper]{IEEEAerospaceCLS}  %

\usepackage[]{graphicx}    %
\newcommand{\ignore}[1]{}  %

\usepackage{graphics} %
\usepackage{graphicx}

\usepackage{subfig}
\usepackage{graphicx}
\usepackage{epsfig} %
\usepackage{amsmath} %
\usepackage{amssymb}  %

\usepackage{bm}
\usepackage{mathtools}

\usepackage{siunitx}
\usepackage{algorithm}
\usepackage[noend]{algpseudocode} %
\algnewcommand\AAND{\textbf{ and }}
\algnewcommand\Or{\textbf{ or }}

\usepackage{color}
\usepackage{flushend}
\usepackage{url}
\usepackage[breaklinks]{hyperref}
\usepackage[capitalise]{cleveref}
\usepackage{booktabs}
\usepackage{makecell}
\usepackage{multirow}

\usepackage[nolist,nohyperlinks]{acronym}
\acrodef{method}[ROAMER]{Robust Off-road Autonomy using Multi-modal State Estimation with Radar Velocity Integration}
\acrodef{gnss}[GNSS]{Global Navigation Satellite System}
\acrodef{icp}[ICP]{Iterative Closest Point}
\acrodef{uas}[UAS]{Unmanned Aerial Systems}
\acrodef{ransac}[RANSAC]{Random Sample Consensus}
\acrodef{slam}[SLAM]{Simultaneous Localization And Mapping}
\acrodef{fmcw}[FMCW]{Frequency Modulated Continuous Wave}
\acrodef{pca}[PCA]{Principal Component Analysis}
\acrodef{ekf}[EKF]{Extended Kalman Filter}
\acrodef{isam2}[iSAM2]{Incremental Smoothing and Mapping}
\acrodef{rio}[RIO]{Radar Inertial Odometry}
\acrodef{rmse}[RMSE]{Root Mean Square Error} 
\acrodef{ape}[APE]{Absolute Pose Error}
\acrodef{cfar}[CFAR]{Constant False Alarm Rate}
\acrodef{snr}[SNR]{Signal to Noise Ratio}
\acrodef{rcs}[RCS]{Radar Cross Section}
\acrodef{imu}[IMU]{Inertial Measurement Unit}
\acrodef{rdmap}[RD-Map]{Range-Doppler Map}
\acrodef{cacfar}[CA-CFAR]{Cell-Averaging \ac{cfar}}
\acrodef{cut}[CUT]{Cell Under Test}
\acrodef{jpl}[JPL]{Jet Propulsion Laboratory}
\acrodef{fft}[FFT]{Fast-Fourier Transform}

\DeclareMathAlphabet{\pazocal}{OMS}{zplm}{m}{n}

\DeclareMathOperator*{\argmin}{argmin}
\DeclareMathOperator*{\argmax}{argmax} %

\DeclareMathAlphabet{\mathpzc}{OT1}{pzc}{m}{it}

\usepackage{array}
\newcolumntype{C}[1]{>{\centering\arraybackslash}p{#1}}
\newcolumntype{M}[1]{>{\raggedright\arraybackslash}p{#1}}

\usepackage{array} 
\newcolumntype{L}[1]{>{\raggedright\let\newline\\\arraybackslash\hspace{0pt}}m{#1}}	
\newcolumntype{S}[1]{>{\centering\let\newline\\\arraybackslash\hspace{0pt}}m{#1}}
\newcolumntype{R}[1]{>{\raggedleft\let\newline\\\arraybackslash\hspace{0pt}}m{#1}}

\makeatletter
\renewcommand*{\@opargbegintheorem}[3]{\trivlist
  \item[\hskip \labelsep{\itshape #1\ #2}] \textit{(#3)}\ }
\makeatother

\usepackage{lipsum}

\begin{document}
\title{ROAMER: Robust Offroad Autonomy using Multimodal State Estimation with Radar Velocity Integration}

\author{%
Morten~Nissov\\
Department of Engineering Cybernetics\\
Norwegian University of Science and Technology\\
Trondheim, 7034, Norway\\
morten.nissov@ntnu.no\\
\and
Shehryar~Khattak\\
Jet Propulsion Laboratory\\
California Institute of Technology, USA\\
Pasadena, CA 91019\\
skhattak@jpl.nasa.gov\\
\and
Jeffrey~A.~Edlund\\
Jet Propulsion Laboratory\\
California Institute of Technology, USA\\
Pasadena, CA 91019\\
jeffrey.a.edlund@jpl.nasa.gov\\
\and
Curtis~Padgett\\
Jet Propulsion Laboratory\\
California Institute of Technology, USA\\
Pasadena, CA 91019\\
curtis.w.padgett@jpl.nasa.gov\\
\and
Kostas~Alexis\\
Department of Engineering Cybernetics\\
Norwegian University of Science and Technology\\
Trondheim, 7034, Norway\\
konstantinos.alexis@ntnu.no
\and
Patrick~Spieler\\
Jet Propulsion Laboratory\\
California Institute of Technology, USA\\
Pasadena, CA 91019\\
patrick.spieler@jpl.nasa.gov\\
\thanks{\footnotesize 979-8-3503-0462-6/24/$\$31.00$ \copyright2024 IEEE}              %
}

\maketitle

\thispagestyle{plain}
\pagestyle{plain}

\maketitle

\thispagestyle{plain}
\pagestyle{plain}

\begin{abstract}
Reliable offroad autonomy requires low-latency, high-accuracy state estimates of pose as well as velocity, which remain viable throughout environments with sub-optimal operating conditions for the utilized perception modalities. As state estimation remains a single point of failure system in the majority of aspiring autonomous systems, failing to address the environmental degradation the perception sensors could potentially experience given the operating conditions, can be a mission-critical shortcoming. In this work, a method for integration of radar velocity information in a LiDAR-inertial odometry solution is proposed, enabling consistent estimation performance even with degraded LiDAR-inertial odometry. The proposed method utilizes the direct velocity-measuring capabilities of an Frequency Modulated Continuous Wave (FMCW) radar sensor to enhance the LiDAR-inertial smoother solution onboard the vehicle through integration of the forward velocity measurement into the graph-based smoother. This leads to increased robustness in the overall estimation solution, even in the absence of LiDAR data. This method was validated by hardware experiments conducted onboard an all-terrain vehicle traveling at high speed, $\SI{\sim 12}{\meter\per\second}$, in demanding offroad environments.
\end{abstract}

\tableofcontents

\section{Introduction}
\begin{figure}[t!]
    \centering
    \includegraphics[width=\linewidth]{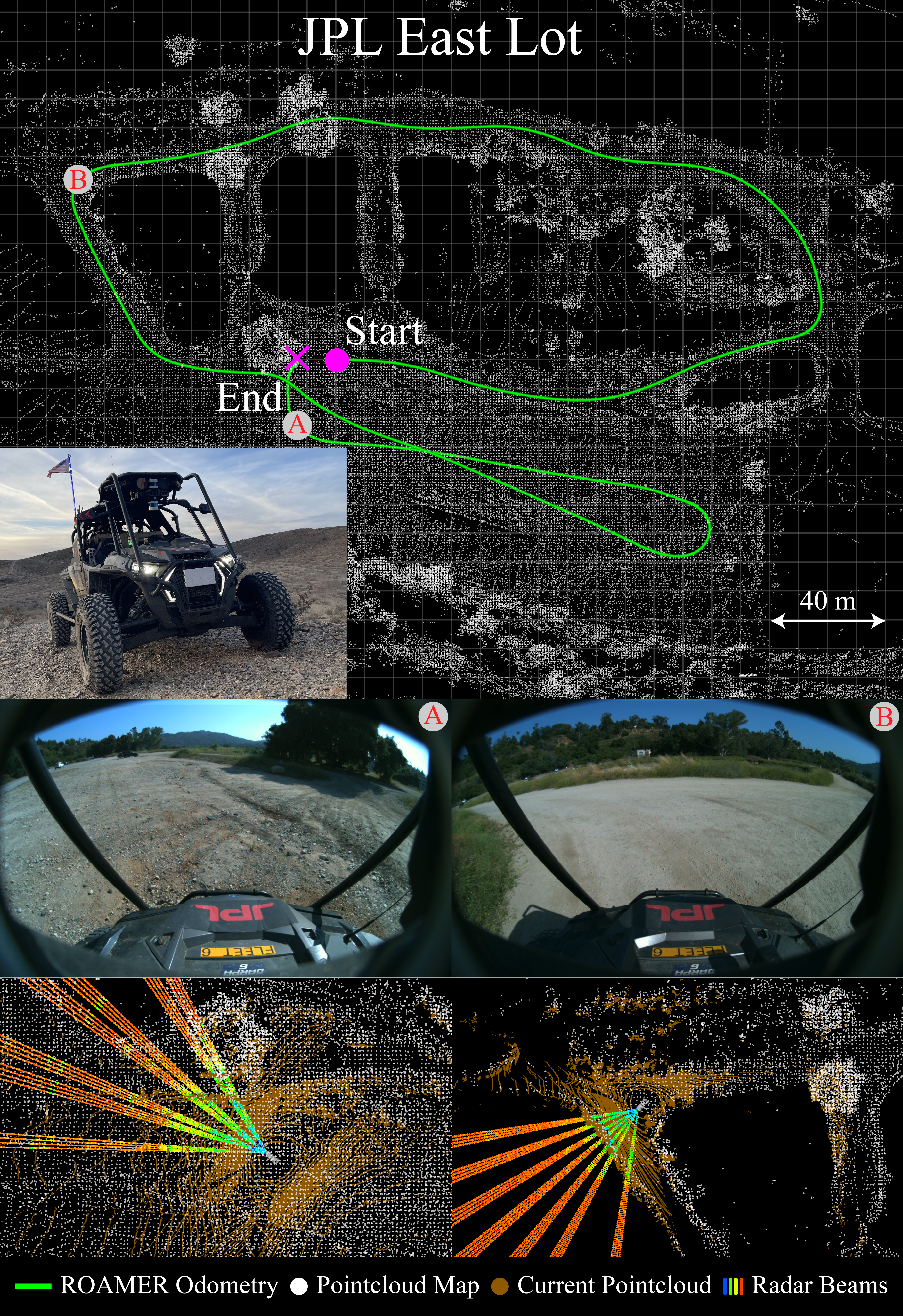}
    \vspace{-4ex}
    \caption{RACER robot platform used to conduct offroad autonomy experiments. Figure includes insets from onboard cameras as well as LiDAR map and radar data visualizations at choice locations. Note, the typical environment in the background, consisting of unmaintained dirt paths, arid conditions, and sparse vegetation.}
    \label{fig:intro}\vspace{-6ex}
\end{figure}

Autonomous robotics navigation in offroad and unstructured environments remains a challenging task for ground-based systems. Although with the increase of interest towards autonomous driving urban vehicles have made significant progress in terms of their autonomy, navigation in offroad settings remains an open problem due to lack of availability of prior knowledge, complexity of terrain structure, and relaxed driving rules. Furthermore, the complexity of the problem increases when traversal is required to be made at high speeds dictated either by the nature of the terrain (e.g. maneuvering steep slopes), or by the urgency of the task (e.g. reaching a plane crash site in the middle of a desert). Robotic high-speed navigation of offroad environments remains an area of high research interest due to its potential to enable fast navigation through unknown and all-terrain environments, not only on Earth but also for planetary science missions, for which reliable rover autonomy can significantly increase the return towards mission science objectives~\cite{highSpeedRover,longRover}. 

Towards realizing robot autonomy in unstructured environments at high speeds, one key research area is enabling robust low-latency state estimation and localization for navigation. To remain robust, most current approaches rely on multimodal sensor fusion to overcome challenges of rapid illumination changes, lack of informative structure, and presence of airborne obscurants such as dust. However, most current loosely-coupled sensor fusion approaches focus on the robust fusion of pose estimates from varying sources and only estimate vehicle linear velocity as an afterthought, i.e linear velocity is only estimated as part of the state vector without provision of any direct velocity measurements being integrated. Nevertheless, accurate estimation of linear velocity remains an important task as most control approaches rely on velocity estimates to provide steering and braking commands, a task that becomes critical for high-speed navigation. Furthermore, as opposed to vehicle rotational velocity, which can be directly measured by utilizing IMU gyroscope measurements, linear velocity updates can only be obtained through the integration of acceleration measurements, making them susceptible to integration errors.

Motivated by the discussion above, this work presents ROAMER, a framework to enable \underline{R}obust \underline{O}ffroad \underline{A}utonomy using \underline{M}ultimodal state \underline{E}stimation with \underline{R}adar velocity integration. Specifically, the proposed method utilizes \ac{fmcw} RADAR measurements to estimate the vehicle's forward velocity and integrate it into a multimodal sensor fusion scheme as a forward velocity measurement update. 
We experimentally demonstrate that the inclusion of these forward velocity updates makes the overall pose estimation process more robust. 
Furthermore, we demonstrate the potential of the proposed method in real world applications by conducting hardware experiments using an all-terrain vehicle tailored for autonomous navigation in complex and unstructured offroad environments at high speeds. An instance of an experiment is shown in~\cref{fig:intro}.

\section{Related Work}

State estimation is a widely studied topic as it is among the core capabilities necessary for enabling autonomous robot operations. Thus, it is required that the estimation machinery is able to function despite challenges present in a given environment. Typically, the states to be estimated include pose and linear velocity, as such aided inertial navigation is a clear solution with variations in the aiding sensor depending on the mission context. Typical aiding sensors include GNSS~\cite{wen2021GNSS}, vision~\cite{geneva2020OpenVINS}, and LiDAR~\cite{fastlio2} which work well in favorable conditions but can perform poorly when challenged. Thermal and radar sensors have become topics of research interest due to their ability to perform in conditions which could prove more challenging for the aforementioned modalities~\cite{khattak2020thermal,doer2021x}.

For high speed ground-based applications, such at autonomous driving, LiDAR seems the obvious choice for its high accuracy and large field of view (FoV). %
LiDAR odometry and mapping methods have played an increasingly important role in both robotic and automotive sectors~\cite{levinson2011AutoDriving}. A large body of work inherits the features and structure described in the seminal paper~\cite{Zhang-RSS-14} which separated out the problem of LiDAR based estimation into high rate velocity estimation and low-rate mapping updates. Other works~\cite{compslam,liosam2020shan,fastlio2} integrate \ac{imu} sensors due to their high frequency measurements in comparison to the slower rates of mechanically spinning LiDARs. Despite the advances in high accuracy Simultaneous Localization And Mapping (SLAM) from LiDAR-based methods, complex environments~\cite{xicp,ebadi2022present} and typical driving conditions~\cite{burnett2022we,sezgin2023safe} and still pose considerable challenges, hence the interest in another common automotive sensing modality: radars.

Radar-based methods have gained traction in both robotics and automotive industries, primarily due to the unique capabilities of these sensors. The survey~\cite{harlow2023new} gives an overview of the broad landscape for advances in \si{\milli\meter} wavelength radars for robotic applications. Seminal works on ego-motion estimation in an automotive domain include \cite{kellner_instantaneous_2013,kellner2014ego}. Many works have been focused on applications in aerial robotics with EKF-based loosely coupled approaches~
\cite{doer2021x}
, tightly-coupled approaches~\cite{michalczyk2022tightly}, as well as graph-based optimization methods~\cite{kramer2021fog}. Specific to the automotive domain there exist works presenting odometry solutions~\cite{burnett2021radar} as well as SLAM solutions~\cite{hong2020radarslam}. In addition to some works investigating the performance potential for radars as a LiDAR replacement~\cite{adolfsson2022lidar}.

However, for autonomous driving at high speed in non-urban environments, when the environment is less predictable and the motion of the vehicle can be aggressive, difficult situations are likely to occur. As a result, instead of considering a single of these modalities it can be necessary to combine their strengths in a fusion approach. Loosely-coupled fusion of different odometry sources was explored in~\cite{brommer2020,nubert2022graph,fakoorian2022rose}. Targeted approaches for combining radar and LiDAR scans were developed in~\cite{fritsche2018fusing,park2019radar}, the former considering landmarks between the modalities and the latter registering radar scans to previously built LiDAR maps. Similarly, the authors in~\cite{yin2022Rall} propose matching radar scans to a LiDAR map utilizing a deep learning-based approach.

Motivated by the discussion above, this work proposes  to fuse forward velocity estimates from radar with accurate pose estimates from LiDARs and low-latency measurements from inertial sensors in a multimodal fusion approach to demonstrate improved state estimation robustness, while maintaining low-latency and accuracy. The proposed approach demonstrates resilient state estimation performance by providing consistent state estimates even during periods of sensor (LiDAR) failure, with no negative impact on estimation performance when compared to normal operating conditions.

\section{Proposed Approach}

\subsection{\textbf{Notation and Coordinate Frames}}
The coordinate frames used in this work are the world frame ($\mathtt{W}$), \ac{imu} frame ($\mathtt{I}$), radar frame ($\mathtt{R}$), and LiDAR frame ($\mathtt{L}$).

A scalar, vector, and matrix variable is denoted by $x$, $\bm{x}$, and $\mathbf{x}$ respectively. Let the homogeneous transformation from $\mathtt{A}$ to $\mathtt{B}$ be ${}_{\mathtt{B}}\mathbf{T}_{\mathtt{A}} \in {\mathrm{SE(3)}}$ which is made of the position of frame $\mathtt{A}$ expressed with respect to frame ${\mathtt{B}}$ (${}_{\mathtt{B}} \bm{p}_{\mathtt{A}} \in{\mathbb{R}}^{3}$) and the rotation from $\mathtt{A}$ to $\mathtt{B}$ (${}_{\mathtt{B}} \mathbf{R}_{\mathtt{A}} \in {\mathrm{SO(3)}}$). Note also, the wedge operator $\bm{x}^{\wedge}$ denotes the skew symmetric matrix of an arbitrary vector $\bm{x}\in{\mathbb{R}}^{3}$.

\subsection{\textbf{Radar Signal Processing}}\label{sec:approach:radar}
\begin{figure}[h!]
    \centering
    \includegraphics[width=\linewidth]{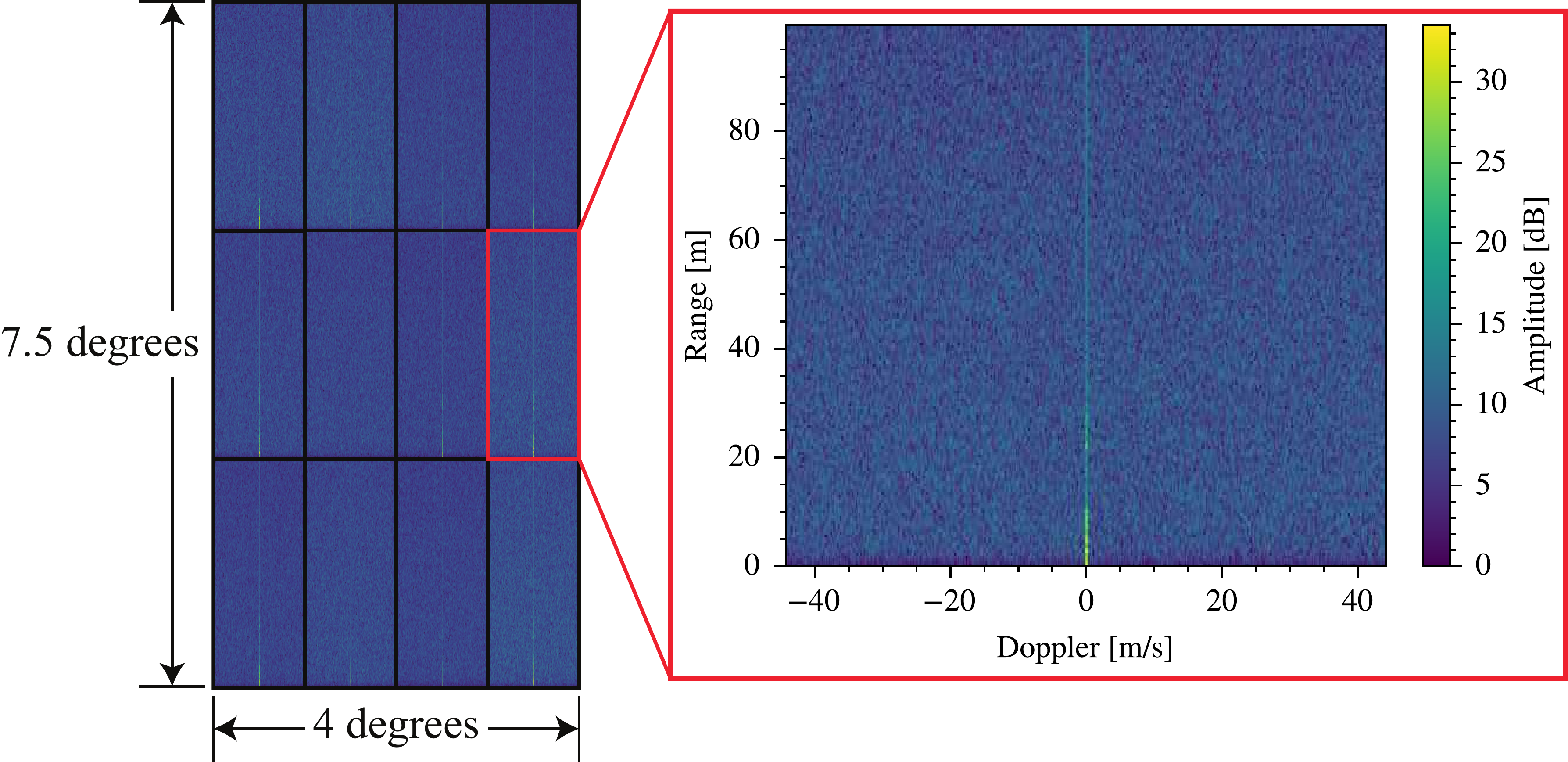}
    \caption{Visualization of a radar beam, which consists of 12 pixels across \SI{4}{\degree} of azimuth and \SI{7.5}{\degree} of elevation. Each pixel consists of a range-doppler image of signal intensities, with size depending on the waveform used. An example of one pixels' range-doppler map can be seen, note this is of a static environment, hence the high intensity returns along 0 doppler.}
    \label{fig:approach:rdmap}
\end{figure}

The proposed method utilizes an \ac{fmcw} radar for ego-velocity estimation. The radar sensor used in this work, the Echodyne EchoDrive, returns a measurement which consists of a single \emph{beam} which itself is made of 12 \emph{pixels}, \SI{1}{\degree} wide and \SI{2.5}{\degree} tall, arranged in three rows and four columns. Each pixel consists of an image-like structure called the \ac{rdmap}. The \ac{rdmap}, as seen in \cref{fig:approach:rdmap}, is a 2D array of signal intensities corresponding to different combinations of range (rows) and doppler (columns) with the size given by the programmed configuration of the radar beam waveform used for a given measurement. This size is defined by known values for the minimum and maximum range and doppler, as well as the range and doppler resolutions.

This data structure is the result of processing both the fast-time and slow-time FFTs over the raw data, as such a high value in one cell corresponds to a high signal intensity, and therefore high likelihood, for an object at that given range-doppler combination.
Each beam measuring period is short (less than \SI{20}{\milli\second}) albeit with a narrow FoV, therefore to cover a larger FoV, radar beams can be sequentially measured along a desired spatial pattern. This pattern is designed such that the azimuth-elevation coverage is acceptable both in terms of resulting total FoV as well as temporal ordering of the particular beams.

\begin{figure}[h!]
    \centering
    \includegraphics[width=\linewidth]{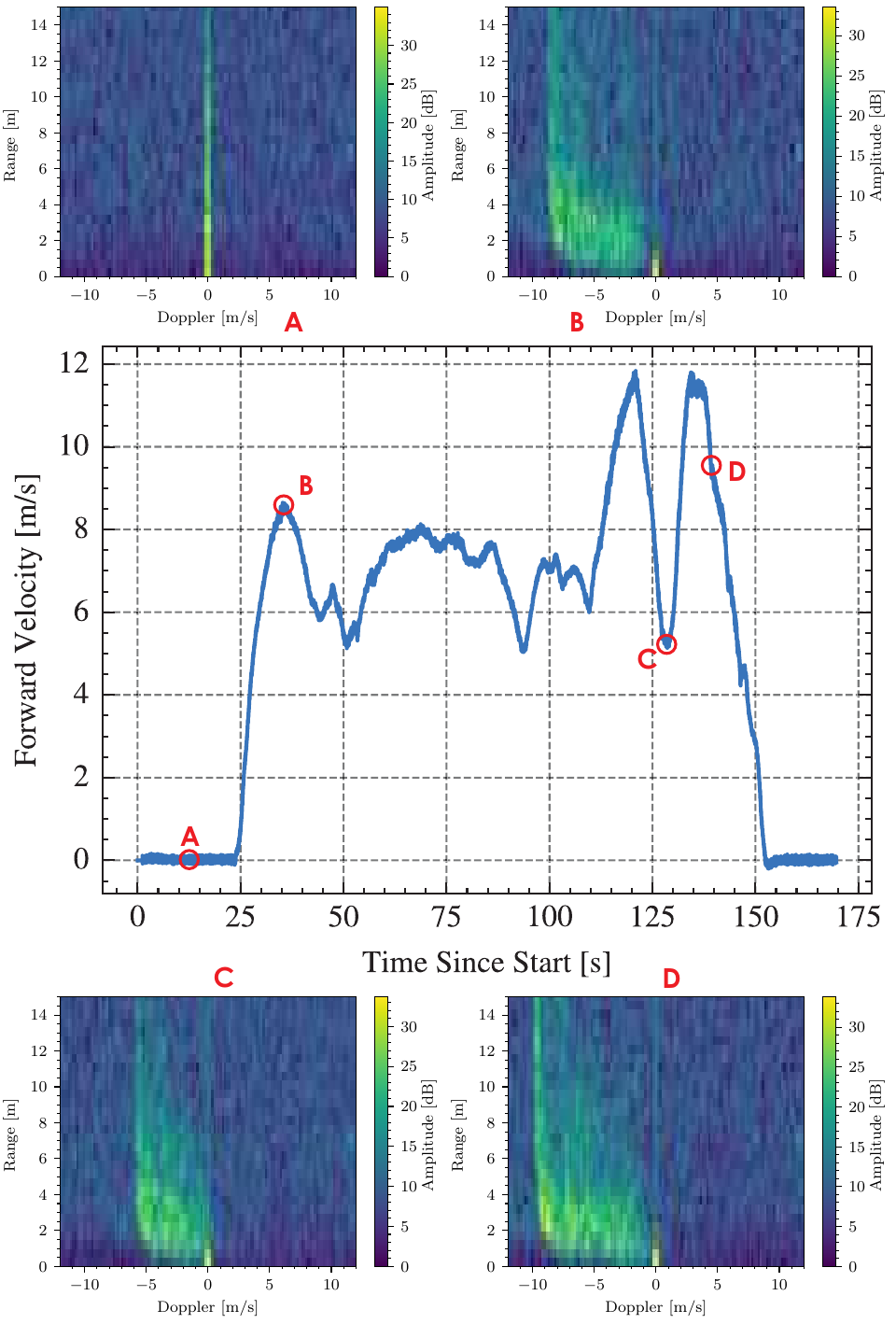}
    \caption{Visualization of a selection of radar beam pixels corresponding to particular points in the velocity profile. These pixels are all along -0.5\textdegree~azimuth and -2.5\textdegree~elevation so they should correspond very closely with forward velocity. Note the quantity of noise in the data but also how the left-most part of the high-intensity \emph{smear} corresponds well to forward velocity. Forward velocity here is taken from the LiDAR-inertial odometry onboard the vehicle. Range-doppler images are cropped in size to more easily see the data.}\label{fig:approach:forward_velocity_with_radar_image}
    \vspace{-2ex}
\end{figure}

For the sensor and waveform used, the \ac{rdmap} is composed of 256 rows by 512 columns of range-doppler measurements, however, only a small number of those range-doppler combinations are likely to correspond to valid targets, therefore filtering is required to extract useful information~\cite{richards2013Fundamentals}.    
The difficulty in separating noise from signal is clear in \cref{fig:approach:forward_velocity_with_radar_image}, which shows the radar images from a pixel pointing along \SI{-0.5}{\degree} azimuth and \SI{-2.5}{\degree} elevation alongside the estimated forward velocity. Clearly there is correlation between forward velocity and the range-doppler images, but it is difficult to discern precisely which range-doppler cells correspond to real targets, as opposed to which are purely noise. Furthermore without a priori knowledge of what the forward velocity should be, this decoding of the image becomes more difficult. As a result, this further motivates the necessity for filtering.

\subsubsection{\textbf{Filtering}}
The techniques and methodologies described in this section come from~\cite[ch. 6]{richards2013Fundamentals} and are summarized here for completeness.

The goal of the filtering is, from a range-doppler image, to classify every cell in the image as one of two classes, where these classes form the basis for our hypothesis test:
\begin{enumerate}
    \item ${\mathcal{H}}_{0}$ The specific cell is a result of interference.
    \item ${\mathcal{H}}_{1}$ The specific cell is a result of interference and reflections from a target.
\end{enumerate}
Note, here the interference can be thought of as an underlying, non-zero noise process with some power $\sigma_w^2$ that varies with respect to many environmental and scenario-based factors. Typically, it is assumed that such interference is Gaussian, for convenience, but it is accepted that this value is non-constant. As such, having an adaptive filtering method is important for consistent performance, which leads to the common usage of \ac{cfar} methods for filtering such range-doppler images.

\ac{cfar} detection is a set of methods which are designed to provide predictable behavior, in terms of detection and false alarm, in more realistic interference conditions. The basic idea is that instead of assuming a constant value for the interference noise power these techniques aim to estimate the local interference levels online and test the radar signal for significant deviations above this noise floor to conclude a valid target, i.e. ${\mathcal{H}}_1$. There exists different methodologies for how this value can be estimated, but in this work \ac{cacfar} is used.

\textbf{\acl{cacfar}}
This method estimates the noise level for a single cell, known as the \ac{cut}, at a time for every cell, after which hypothesis testing can be performed over the entire image. The hypothesis test in this case checks if the \ac{cut} signal exceeds a threshold, if true it is classified as ${\mathcal{H}}_1$ and if false as ${\mathcal{H}}_0$. The threshold is determined as a function of the noise estimate, where the desired probability of false alarm $P_{FA}$ determines how much above the noise estimate the threshold must be set to guarantee this $P_{FA}$. This process of hypothesis testing continues through all possible cells until all cells of the range-doppler image have been classified. Doing so requires knowledge of the interference power in the \ac{cut}, which must be estimated from data due to the difficulties mentioned previously. The estimation of this value makes two assumptions regarding the neighboring cells to the \ac{cut}~\cite{richards2013Fundamentals}
:
\begin{enumerate}\label{enum:cacfar_assumptions}
    \item That the interference of the neighboring cells have the same statistical characteristics as the \ac{cut}, essentially that the cells are i.i.d.
    \item That the neighboring cells contain exclusively interference, e.g. no reflections from any targets.
\end{enumerate}

With these satisfied, the interference power $\sigma_w^2$ can be approximated as the average of the signal intensity in the neighboring region and the corresponding threshold $T$ for hypothesis testing can be set such that
\begin{equation}
    T = \alpha \sigma_w^2
\end{equation}
where $\alpha$ is the scaling parameter determined by the desired $P_{FA}$. Specifically for \ac{cacfar}, $\alpha$ can be calculated according as
\begin{equation}\label{eq:approach:cfar:alpha}
    \alpha = N \left(P_{FA}^{-\tfrac{1}{N}} - 1 \right)
\end{equation}
where $N$ is the number of cells used for the noise estimate.

\begin{figure}[h!]
    \centering
    \includegraphics[width=\linewidth]{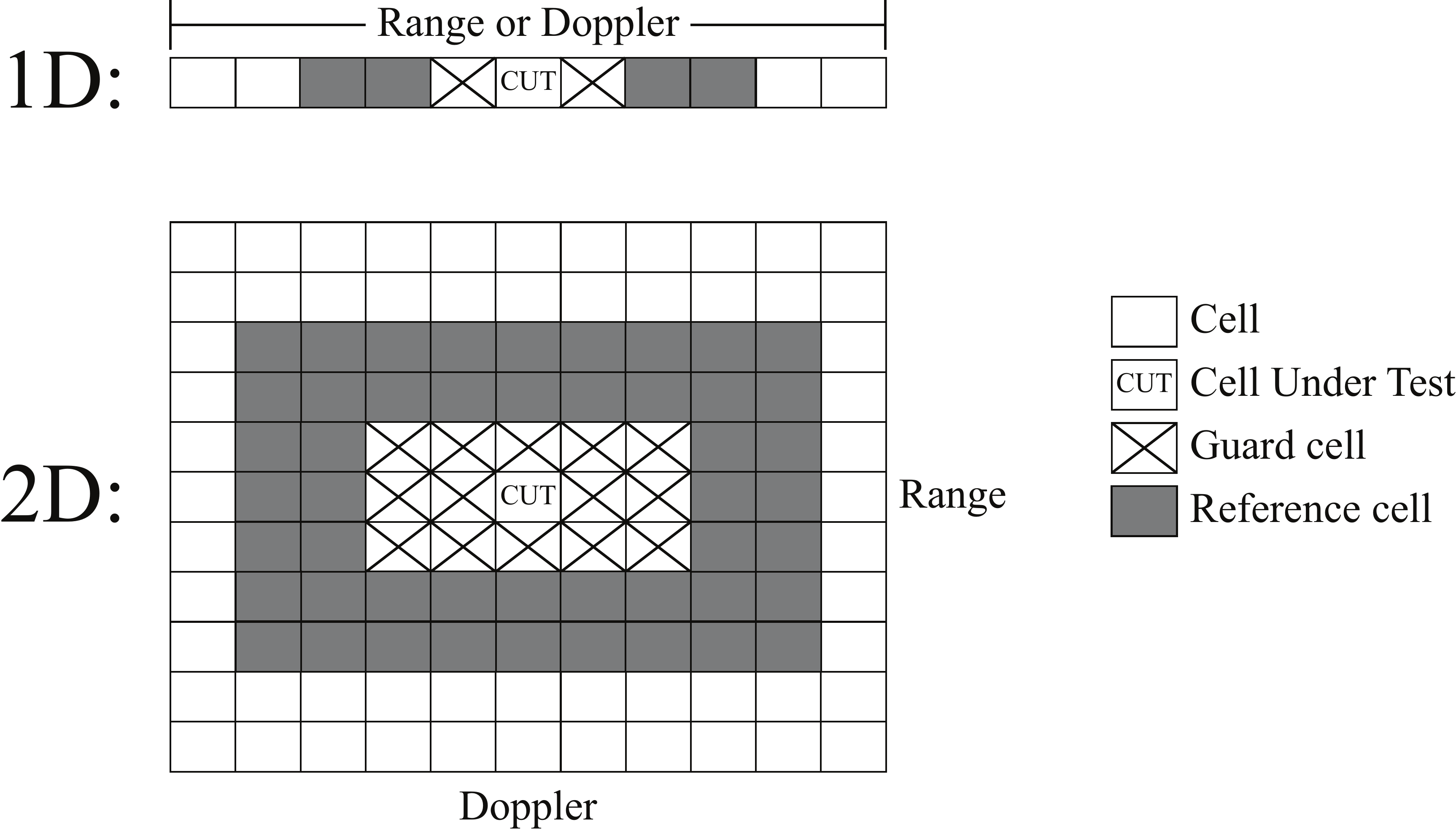}
    \caption{Redraw of the visualization in \protect\cite[Figure 6.19]{richards2013Fundamentals}, depicting 1D and 2D \ac{cfar} kernels used for estimating the interference noise of the \ac{cut}.}
    \label{fig:approach:cfar_window}
\end{figure}

\Cref{fig:approach:cfar_window} shows a visualization of how this averaging can be accomplished in both 1D and 2D cases, as the calculation is more complicated than a simple average. The \ac{cfar} \emph{window} here is made up of two types of cells, reference and guard cells. Guard cells sit immediately beside the \ac{cut} and act as a buffer, their purpose is that if the target in the \ac{cut} straddles over several cells then the guard cells prevent this from affecting the interference estimate. Reference cells are simply those used for estimating the interference power, and are placed beside the guard cells. Note, that the example in \cref{fig:approach:cfar_window} visualizes an arbitrary number of guard and reference cells.

\subsubsection{\textbf{Doppler Estimate from Radar Beam}}
Given a static environment, ego-motion results in a relative velocity between the autonomous agent and the objects in the environment. This fact is what enables radar sensors to be used for ego-velocity measurement and estimation. \begin{figure}[h!]
    \centering
    \includegraphics[width=\linewidth]{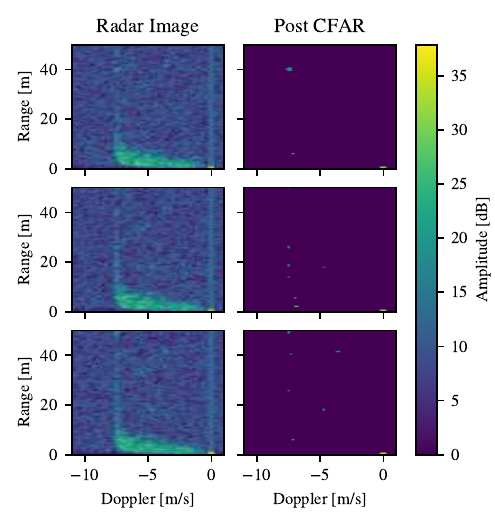}
    \vspace{-4ex}
    \caption{Visualization of the range-doppler image before and after applying the \ac{cacfar} filter, where the right column represents all cells in the range-doppler image from the left column which were classified as targets (${\mathcal{H}}_{1}$) rather than interference. Note, that not all images will necessarily have post filter targets in the same doppler column, motivating the usage of consensus to reduce the likelihood of outliers in the radial speed measurement.}
    \label{fig:approach:radar_filtering}
\end{figure}
\begin{figure*}[h!]
    \centering
    \includegraphics[width=\linewidth]{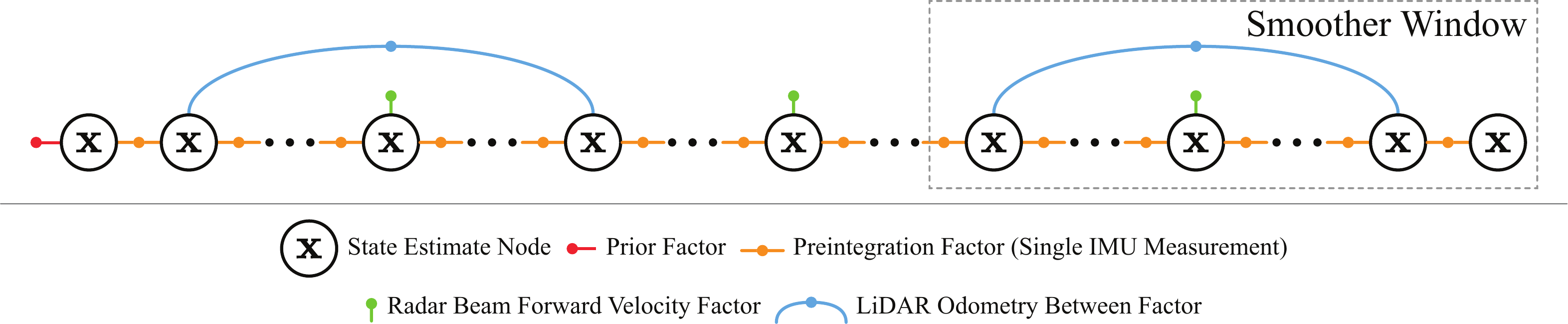}
    \caption{Architecture of the factor graph, including factors representing the information from \ac{imu}, LiDAR, and radar measurements. Note, a new graph node is created upon receiving each \ac{imu} measurement, and other sensor measurements are fixed to the nearest graph node by time.}
    \label{fig:graph_architecture}
\end{figure*}
The actual measured quantity from the radar is the radial speed, also known as doppler, and this quantity is related to this relative motion by
\begin{equation}\label{eq:approach:doppler_relationship}
    v_d = -\bm{r}^{\top} {}_{\mathtt{R}}\bm{v}
\end{equation}
where $v_d$ is the radial speed along bearing vector $\bm{r}$ resulting from ego-velocity expressed in the radar-frame ${}_{\mathtt{R}} \bm{v}$. As this relationship, between doppler and ego-velocity, is not a function of the range of a given measurement, the signal processing is designed to estimate the most likely radial speed without regard for the range of the targets.

Upon receiving a radar beam measurement, each pixel is filtered using the \ac{cacfar} method described previously. Next, each filtered \ac{rdmap} is trimmed in size corresponding to what are considered to be reasonable minimum and maximum bounds for the sensor detection range and operational velocity envelope of the vehicle. Afterwards, we try to find the most probable value for doppler by looking for the column with the greatest number of non-zero cells, which results in a single value for each pixel. To reduce outliers we check for consensus among the pixels, in this case by calculating the mode of the distribution, and enforcing a minimum number of votes for a given velocity value to constitute as a valid measurement. This step is important as, even after filtering with \ac{cacfar}, some interference will still be classified as targets, see \cref{fig:approach:radar_filtering} for inconsistencies between adjacent pixels at \SI{-2.5}{\degree}. It is unlikely this happens in the same way for all pixels, as a result performing a consensus step increases the robustness of the radial speed measurement.

After the aforementioned steps, a beam consisting of 12 \acp{rdmap} is condensed into a single doppler measurement $v_d$ directed at the beam's center azimuth and elevation angles.

\subsection{\textbf{State Estimation using Factor Graphs}}
The state space for this formulation consists of the transform from the \ac{imu} frame to the world frame ${}_{\mathtt{W}}{\mathbf{T}}_{\mathtt{I}} \in {\mathrm{SE(3)}}$ (consisting of translation ${}_{\mathtt{W}}\bm{p}_{\mathtt{I}}$ and rotation ${}_{\mathtt{W}}{\mathbf{R}}_{\mathtt{I}} \in {\mathrm{SO(3)}}$), linear velocity of the \ac{imu} with respect to $\mathtt{W}$ expressed in world frame ${}_{\mathtt{W}}\bm{v}_{\mathtt{WI}}$, and \ac{imu} biases ${}_{\mathtt{I}}\bm{b}\in{\mathbb{R}}^6$ (accelerometer $\bm{b}_{a}$ and gyroscope $\bm{b}_{g}$ terms), resulting in the state vector
\begin{equation}
    \mathbf{x} = \begin{bmatrix}
        {}_{\mathtt{W}}\mathbf{T}_{\mathtt{I}} &{}_{\mathtt{W}}\bm{v}_{\mathtt{WI}} &{}_{\mathtt{I}}\bm{b}
    \end{bmatrix}
\end{equation}
The proposed method is using the graph architecture described in \cite{nubert2022graph} to solve the MAP optimization with a windowed smoother. Generally, this can consist of many sensors providing information, but the existing estimation onboard the vehicle uses \ac{imu} and either LiDAR pose unary or between factors, to which the proposed method adds radar for increased robustness.

Let the \ac{imu} measurement at time $i$ be ${\mathcal{I}}_i$, the LiDAR measurement at time $j$ be ${\mathcal{L}}_j$, and the radar measurement at time $k$ be ${\mathcal{R}}_k$. Where the set of all \ac{imu}, LiDAR, and radar measurement times collected up to time $m$ is ${}_{\mathtt{I}}{\mathcal{M}}_m$, ${}_{\mathtt{L}}{\mathcal{M}}_m$, and ${}_{\mathtt{R}}{\mathcal{M}}_m$ respectively. Thus, the set of all measurements up to time $m$ is 
\begin{equation}
    {\mathcal{Z}}_m=\{{\mathcal{I}}_i, {\mathcal{L}}_j, {\mathcal{R}}_k\}_{i\in{}_{\mathtt{I}}{\mathcal{M}}_m,\ j\in{}_{\mathtt{L}}{\mathcal{M}}_m,\ k\in{}_{\mathtt{R}}{\mathcal{M}}_m}
\end{equation}
The graph thus solves for the MAP estimate for a window of \ac{imu} states given the smoother lag $l$
\begin{multline}
    {\mathcal{X}}_{m-l:m}^* = \argmax_{{\mathcal{X}}_{m-l:m}} p\left( {\mathcal{X}}_{m-l:m} \vert {\mathcal{Z}}_{m-l:m} \right) \\\propto p\left( {\mathcal{X}}_{0} \right) p\left( {\mathcal{Z}}_{m-l:m} \vert {\mathcal{X}}_{m-l:m} \right)
\end{multline}
where ${\mathcal{X}}_{m-l:m}$ is the windowed set of states from $m-l$ to $m$ and $m\in {}_{\mathit{I}}{\mathcal{M}}$.
Assuming non-zero mean, Gaussian noise models this can be rewritten as a sum of weighted, squared residual errors
\begin{multline}
     {\mathcal{X}}_{m-l:m}^* = \argmin_{ {\mathcal{X}}_{m-l:m}} \Big( \lVert \bm{e}_{0} \rVert_{\bm\Sigma_0}^2 + \Sigma_{i\in{}_{\mathtt{I}}{\mathcal{M}}_{m-l:m}} \lVert \bm{e}_{{\mathcal{I}}_{i}} \rVert_{\bm\Sigma_{\mathcal{I}}}^2\\ + \Sigma_{j\in{}_{\mathtt{L}}{\mathcal{M}}_{m-l:m}}\lVert \bm{e}_{{\mathcal{L}}_j} \rVert_{\bm\Sigma_{\mathcal{L}}}^2 + \Sigma_{k\in{}_{\mathtt{R}}{\mathcal{M}}_{m-l:m}}\lVert \bm{e}_{{\mathcal{R}}_k} \rVert_{\bm\Sigma_{\mathcal{R}}}^2 \Big)
\end{multline}
where $\bm{e}_{0}$, $\bm{e}_{{\mathcal{I}}_{i}}$, $\bm{e}_{{\mathcal{L}}_j}$, and $\bm{e}_{{\mathcal{R}}_k}$ are the residual errors for the marginalization prior and subsequent measurements and $\bm\Sigma_0$, $\bm\Sigma_{\mathcal{I}}$, $\bm\Sigma_{\mathcal{L}}$, and $\bm\Sigma_{\mathcal{R}}$ are the corresponding covariance matrices. The residual errors are functions of the state space which quantify discrepancies between the measurements and measurement predictions.

There are three types of factors, derived from the sensor measurements, used in the factor graph: one for \ac{imu}, one for LiDAR, and one for radar. In the following sections, the way the particular measurements are included in the graph estimator will be described. The overall architecture is shown in \cref{fig:graph_architecture}.
Note, all Jacobians are derived analytically and implemented as such, this is done because it provides significant cost savings in terms of computation. Derivations are not shown here, but an introduction to calculus with Lie groups can be found in~\cite{solà2021micro}.

\subsubsection{\textbf{IMU Measurements}}
In order to create high frequency output for use in control, the method proposed in \cite{nubert2022graph} creates a node in the graph for each \ac{imu} measurement, connected with a pre-integrated \ac{imu} factor with the residuals and covariance as described in~\cite{forster_-manifold_2017}:
\begin{equation}
    \bm{e}_{\mathcal{I}} = \begin{bmatrix}
        \bm{e}_{\Delta{}_{\mathtt{W}}{\mathbf{R}}_{\mathtt{I}}}^\top &\bm{e}_{\Delta{}_{\mathtt{W}}\bm{v}_{\mathtt{WI}}}^\top &\bm{e}_{\Delta{}_{\mathtt{W}}\bm{p}_{\mathtt{I}}}^\top
    \end{bmatrix}^\top,\quad \bm\Sigma_{\mathcal{I}}
\end{equation}
where $\bm{e}_{\Delta{}_{\mathtt{W}}{\mathbf{R}}_{\mathtt{I}}}$, $\bm{e}_{\Delta{}_{\mathtt{W}}\bm{v}_{\mathtt{WI}}}$, and $\bm{e}_{\Delta{}_{\mathtt{W}}\bm{p}_{\mathtt{I}}}$ are the residuals, of the \ac{imu} factor, with respect to the orientation, velocity, and position. This means that the graph is linked to the \ac{imu} frequency and timestamps, other sensor measurements which do not necessarily align to these timestamps are added to the closest, by timestamp, node in the graph.

\subsubsection{\textbf{LiDAR Measurements}}
The LiDAR measurements are used, as described in~\cite{fakoorian2022rose}, to calculate relative transforms from the previous to current LiDAR measurement. 
Following~\cite{nubert2022graph} the LiDAR odometry can be implemented either as a pseudo-global or relative transformation factor. For smoother estimates from the proposed method (demonstrated in \cref{sec:evaluation}) it will be added as a relative transformation between the current pose ${}_{\mathtt{W}} {\mathbf{T}}_{{\mathtt{I}}_j}$ and previous pose ${}_{\mathtt{W}} {\mathbf{T}}_{{\mathtt{I}}_{j-1}}$ such that
\begin{equation}
    \bm{e}_{{\mathcal{L}}_j} = \log \left( \left( {}_{{\mathtt{I}}_{j-1}} {\mathbf{T}}_{{\mathtt{I}}_{j}} \right)^{-1} \left( {}_{\mathtt{W}}{\mathbf{T}}_{{\mathtt{I}}_{j-1}} \right)^{-1} {}_{\mathtt{W}}{\mathbf{T}}_{{\mathtt{I}}_j} \right)
\end{equation}
where 
${}_{{\mathtt{I}}_{j-1}} {\mathbf{T}}_{{\mathtt{I}}_{j}}$ is the relative transform created from the LiDAR odometry following the approach in \cite{nubert2022graph}, after transforming from $\mathtt{L}$ to $\mathtt{I}$ through the extrinsics,
and $\log$ is the logarithmic mapping from manifold to tangent space of ${\mathrm{SE(3)}}$.

\subsubsection{\textbf{Radar Measurements}}
Given the radial speed from a particular beam, the forward velocity in $\mathtt{R}$-frame from this measurement can be approximated as
\begin{equation}
    {}_{\mathtt{R}}\bm{v}_{\mathtt{WR}}^{x} = -\frac{v_d}{\cos\theta}
\end{equation}
where ${}_{\mathtt{R}}\bm{v}_{\mathtt{WR}}^{x}$ is the forward velocity and $\theta$ is the azimuth angle. Note, this is derived from the relationship introduced in \cref{eq:approach:doppler_relationship} with some simplifying assumption. This approximation assumes that lateral speed and elevation angle are zero, the former of which should be close to true given the vehicle locomotion and the latter of which almost true given the chosen measuring pattern.

To be usable in the graph, this measured quantity needs to be related to a function of the state space variables. This relationship can be derived by calculating the $\mathtt{R}$-frame linear velocity and then transforming and compensating for the rotating reference frame by the \ac{imu}-frame angular velocity ${}_{\mathtt{I}}\bm{\omega}$, such that
\begin{equation}
    {}_{\mathtt{R}}\bm{v}_{\mathtt{WR}} = {}_{\mathtt{R}}{\mathbf{R}}_{\mathtt{I}} \left( {}_{\mathtt{I}}{\mathbf{R}}_{\mathtt{W}} {}_{\mathtt{W}}\bm{v}_{\mathtt{WI}} + {}_{\mathtt{I}}\bm{\omega} \times {}_{\mathtt{I}}\bm{p}_{\mathtt{R}} \right)
\end{equation}
where the radar-\ac{imu} extrinsics are given by $\{{}_{\mathtt{I}}{\mathbf{R}}_{\mathtt{R}}, {}_{\mathtt{I}}\bm{p}_{\mathtt{R}}\}\in {\mathrm{SE(3)}}$.

Isolating the forward velocity from this relationship is simply done by pre-multiplying with a selection vector $\bm{s}^x$ defined as
\begin{equation}
    \bm{s}^x = \begin{bmatrix}
    1 &0 &0
\end{bmatrix}
\end{equation}
As such, the forward velocity measurement ${}_{\mathtt{R}}\tilde{\bm{v}}_{\mathtt{WR}}^{x}$ can be incorporated into the graph, using the \ac{imu} measurement for angular velocity and adding the bias estimate compensation, with the following error function
\begin{equation}
    {e}_{\mathcal{R}} = \bm{s}^x
    {}_{\mathtt{R}}{\mathbf{R}}_{\mathtt{I}} \left( {}_{\mathtt{I}}{\mathbf{R}}_{\mathtt{W}} {}_{\mathtt{W}}{\bm{v}}_{\mathtt{WI}} + \left( {}_{\mathtt{I}}{\bm{\omega}} - {\bm{b}}_{g} \right) \times {}_{\mathtt{I}}\bm{p}_{\mathtt{R}} \right) - {}_{\mathtt{R}}\tilde{\bm{v}}_{\mathtt{WR}}^{x}
\end{equation}
The non-zero Jacobians of the residual with respect to the state space are
\begin{equation}
    \begin{aligned}
        \frac{\partial {e}_{\mathcal{R}}}{\partial {}_{\mathtt{W}} {\mathbf{R}}_{\mathtt{I}}} &= \bm{s}^x {}_{\mathtt{R}}{\mathbf{R}}_{\mathtt{I}} \left( {}_{\mathtt{I}}{\mathbf{R}}_{\mathtt{W}} {}_{\mathtt{W}}\bm{v}_{\mathtt{WI}} \right)^\wedge\\
        \frac{\partial {e}_{\mathcal{R}}}{\partial {}_{\mathtt{W}} \bm{v}_{\mathtt{WI}}} &= \bm{s}^x {}_{\mathtt{R}}{\mathbf{R}}_{\mathtt{I}} {}_{\mathtt{I}}{\mathbf{R}}_{\mathtt{W}}\\
        \frac{\partial {e}_{\mathcal{R}}}{\partial \bm{b}_g} &= \bm{s}^x {}_{\mathtt{R}}{\mathbf{R}}_{\mathtt{I}} \left( {}_{\mathtt{I}}\bm{p}_{\mathtt{R}} \right)^\wedge
    \end{aligned}
\end{equation}

\subsection{\textbf{Initialization}}
The proposed method implements an initialization routine to aid in estimation convergence. This consists of a short (\SI{\sim 1}{\second}), static duration in the beginning where \ac{imu} data is accumulated for estimation of initial roll and pitch orientation, gravity magnitude, and the gyroscope biases.

\subsection{\textbf{Implementation Details}}
The proposed method is implemented in C++ using the open-source libraries GTSAM~\cite{gtsam} and Graph-MSF~\cite{nubert2022graph} for factor graph optimization. For the windowed smooter, we use the implementation provided by GTSAM with a window size of \SI{1.5}{\second}. Furthermore, the LiDAR and radar derived factors are also wrapped with Huber norm robust cost functions to further improve robustness to outliers in the measurements.

\section{Evaluation}\label{sec:evaluation}
For validation the proposed method was tested using a modified Polaris
RZR all-terrain vehicle, shown in~\cref{fig:intro}.
For testing, data from an Xsens MTi-630 \ac{imu} (\SI{400}{\hertz}), three Velodyne VLP-32 LiDARs (\SI{10}{\hertz}), and an Echodyne EchoDrive \ac{fmcw} radar were used.
The radar chirp configuration parameters are presented in \cref{tab:evaluation:radar_chirp}, and a beam sweep covering $\pm$\SI{40}{\degree} in azimuth and \SI{-2.5}{\degree} elevation was used with the idea being that
ground reflections and low height objects will comprise a significant portion of valid static targets.
For ground truth a Piksi Multi GNSS sensor was used.
\begin{table}[h!]
    \centering
    \caption{Parameters for the {S21a} EchoDrive chirp.}
    \label{tab:evaluation:radar_chirp}
    \vspace{-1ex}
    \begin{tabular}{ll}
        \specialrule{.1em}{.05em}{.3em}
        Parameter   &Value\\
        \specialrule{.05em}{0.3em}{0.3em}
        Max Range   &\SI{100}{\meter}\\
        Range Resolution   &\SI{0.49}{\meter}\\
        Max Doppler   &\SI{43.178}{\meter\per\second}\\
        Doppler Resolution   &\SI{0.169}{\meter\per\second}\\
        Beam Sampling Time   &\SI{15.8}{\milli\second}\\
        \specialrule{.1em}{.3em}{.05em}
    \end{tabular}
\end{table}

A high-speed driving dataset was captured in an unstructured environment near NASA \ac{jpl} in Pasadena, USA. This environment is rugged in nature and contains an offroad trail on which the vehicle can emulate driving in back-country environments. During the experiment, starting from a stand still position the vehicle first accelerates forward  driving around a loop on the offroad trail, with small foliage, sparse trees and water on either side. After completing one loop, the vehicle moves to a lower elevation path and re-accelerates. After driving to the edge of the offroad environment, the vehicle makes a sharp turn and returns to the approximate vicinity of the starting location, traversing a total distance of $\SI{\sim 950}{\meter}$ and achieving a top speed of $\SI{\sim 12}{\meter\per\second}$ during the experiment. Ground truth for this experiment comes from a GPS-aided inertial navigation smoother implementation which utilizes \ac{imu} preintegration and creates a local frame from geodetic coordinates such that state estimates can be framed in the local coordinate system.
\subsection{\textbf{Validation Experiment}}
\begin{figure}[h!]
    \centering
    \includegraphics[width=\linewidth]{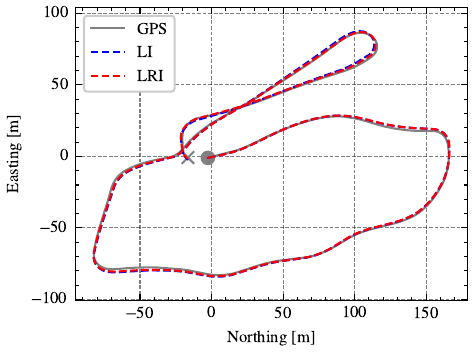}
    \vspace{-4ex}
    \caption{Plot of position estimates from GPS-aided INS ground truth (GPS), LiDAR-inertial odometry (LI), and the proposed LiDAR-radar-inertial odometry (LRI). Note the gray circle and cross mark the start and end of the trajectory, respectively.}
    \label{fig:evaluation:xy_lidarOnly_radarOnly}
    \vspace{-1ex}
\end{figure}

\begin{figure}[h!]
    \centering
    \includegraphics[width=\linewidth]{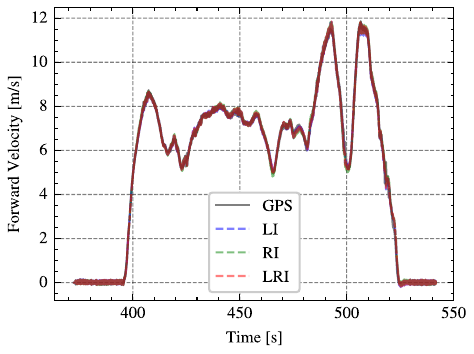}
    \vspace{-4ex}
    \caption{Plot of forward velocity estimates from GPS-aided INS ground truth (GPS), LiDAR-inertial odometry (LI), radar-inertial odometry (RI), and the proposed LiDAR-radar-inertial odometry (LRI).}
    \label{fig:evaluation:vxyz_lidarOnly_radarOnly}
\end{figure}

First, the forward velocity factor implementation is validated through comparison with the onboard LiDAR odometry (LI). Looking at \cref{fig:evaluation:xy_lidarOnly_radarOnly,fig:evaluation:vxyz_lidarOnly_radarOnly} it can be seen that the combined LiDAR-radar-inertial (LRI) system has performance comparable to the existing LI-odometry, see RPE in \cref{tab:evaluation:rpe_10m}, demonstrating that no system-level performance degradation is introduced by the radar velocity integration. 
\begin{table}[h!]
    \centering
    \caption{Transform RPE with \SI{10}{\meter} delta of the different methods in normal operating conditions.}
    \label{tab:evaluation:rpe_10m}
    \vspace{-1ex}
    \begin{tabular}{lcc}
        \specialrule{.1em}{.05em}{.3em}
        Method   &RMSE [-]  &Std. Dev. [-]\\
        \specialrule{.05em}{0.3em}{0.3em}
        LI    &0.342   &0.195\\
        LRI    &{0.348}   &{0.198}\\
        \specialrule{.1em}{.3em}{.05em}
    \end{tabular}
\end{table}

Furthermore, comparing the accuracy of the velocity estimation in \cref{tab:evaluation:vel_err} demonstrates that the velocity estimation from purely RI is slightly worse, but when combined with LiDAR in LRI the result is comparable (different at \SI{<4}{\milli\meter\per\second}) to LI. Therefore inclusion of the radar measurements does not come at a cost to the overall system performance. As such the improvement in robustness can be investigated independently.
\begin{table}[h!]
    \centering
    \caption{RMSE and standard deviation of the error in forward velocity estimate in normal operating conditions.}
    \label{tab:evaluation:vel_err}
    \vspace{-1ex}
    \begin{tabular}{lcc}
        \specialrule{.1em}{.05em}{.3em}
        Method   &RMSE [\si{\meter\per\second}]  &Std. Dev. [\si{\meter\per\second}]\\
        \specialrule{.05em}{0.3em}{0.3em}
        LI    &0.159	&0.159\\
        RI    &0.180	&0.178\\
        LRI   &0.162	&0.162\\
        \specialrule{.1em}{.3em}{.05em}
    \end{tabular}
    \vspace{-2ex}
\end{table}

\subsection{\textbf{Robustness Experiment}}
\begin{figure}[h!]
    \centering
    \includegraphics[width=\linewidth]{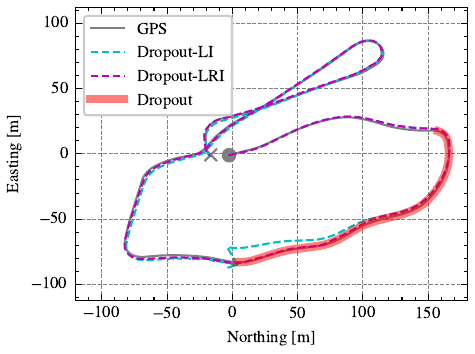}
    \vspace{-4ex}
    \caption{Plot of position estimates from GPS-aided INS ground truth (GPS), LiDAR-inertial odometry (LI), and the proposed LiDAR-radar-inertial odometry (LRI) when dropout is introduced. Note the gray circle and cross mark the start and end of the trajectory, respectively.}
    \label{fig:xy_dropout}
\end{figure}

\begin{figure}[h!]
    \centering
    \includegraphics[width=\linewidth]{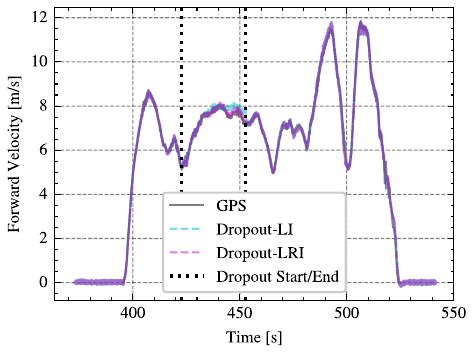}
    \vspace{-4ex}
    \caption{Plot of forward velocity estimates from GPS-aided INS ground truth (GPS), LiDAR-inertial odometry (LI), and the proposed LiDAR-radar-inertial odometry (LRI) when dropout is introduced.}
    \label{fig:vxyz_dropout}
\end{figure}

\begin{figure}[h!]
    \centering
    \includegraphics[width=\linewidth]{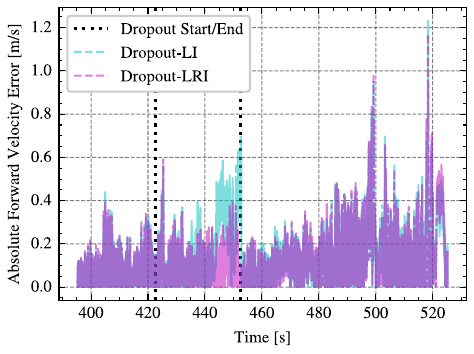}
    \vspace{-4ex}
    \caption{Plot of forward velocity estimate error squared from LiDAR-inertial odometry (LI) as well as the proposed LiDAR-radar-inertial odometry (LRI) in the dropout test. Error is calculated with respect to ground truth.}
    \label{fig:vxyz_error_dropout}
\end{figure}

In order to assess whether the addition of radar information can increase the overall robustness of the state estimator, we intentionally degraded the quality of the LiDAR odometry solution. A measurement dropout was applied to the LiDAR odometry factors being added to the factor graph between \SI{50}{\second} and \SI{80}{\second} from the experiment start. This is meant to emulate the effect of reduced LI quality due to unforeseen circumstances. The immediate effect of the dropout can be seen in position and velocity plots \cref{fig:xy_dropout,fig:vxyz_dropout,fig:vxyz_error_dropout} in the form of drift in position and error in velocity, followed by a large jump to the correct position when LiDAR information is re-introduced to the graph. The degraded performance is also clear in \cref{tab:evaluation:rpe_10m} by the increase in RPE and \cref{tab:evaluation:vel_err} by the increase in velocity error.
\begin{table}[h!]
    \centering
    \caption{Transform RPE with \SI{10}{\meter} delta of the different methods with dropout.}
    \label{tab:evaluation:rpe_10m:dropout}
    \vspace{-1ex}
    \begin{tabular}{lcc}
        \specialrule{.1em}{.05em}{.3em}
        Method   &RMSE [-]  &Std. Dev. [-]\\
        \specialrule{.05em}{0.3em}{0.3em}
        Dropout-LI    &1.661   &1.535\\
        Dropout-LRI   &\textbf{0.405}   &\textbf{0.221}\\
        \specialrule{.1em}{.3em}{.05em}
    \end{tabular}
    \vspace{-2ex}
\end{table}
\begin{table}[h!]
    \centering
    \caption{RMSE and standard deviation of the error in forward velocity estimate with dropout.}
    \label{tab:evaluation:vel_err_dropout}
    \vspace{-1ex}
    \begin{tabular}{lcc}
        \specialrule{.1em}{.05em}{.3em}
        Method   &RMSE [\si{\meter\per\second}]  &Std. Dev. [\si{\meter\per\second}]\\
        \specialrule{.05em}{0.3em}{0.3em}
        Dropout-LI    &0.185	&0.184\\
        Dropout-LRI   &\textbf{0.162}	&\textbf{0.162}\\
        \specialrule{.1em}{.3em}{.05em}
    \end{tabular}
    \vspace{-2ex}
\end{table}

However, with the proposed integration of radar measurements, an improvement in the pose estimates can be seen in~\cref{tab:evaluation:rpe_10m}, as well as an improvement in the linear velocity estimates can be seen in~\cref{fig:xy_dropout,fig:vxyz_dropout}, with the error returning to pre-dropout levels,~\cref{tab:evaluation:vel_err}. It can be noticed that the velocity estimate remains smooth with minimal jump when LiDAR is re-introduced. Furthermore, the differences between LI and LRI velocity errors during the dropout region are clearly noticeable in the velocity error plot in~\cref{fig:vxyz_error_dropout}.

\subsection{\textbf{Limitations}}
\begin{figure}[h!]
    \centering
    \includegraphics[width=\linewidth]{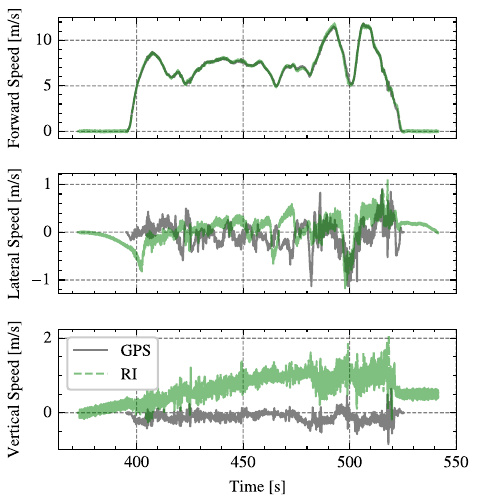}
    \vspace{-4ex}
    \caption{Velocity estimates in body frame from GPS-aided INS (GPS) and radar-inertial odometry (RI).}
    \label{fig:v_all:gps_ri}
    \vspace{-1ex}
\end{figure}

The chosen formulation has some limitations, particularly related to the integration of radar measurements into the graph using only forward velocity factors. Although using each radar beam measurement on arrival incorporates information into the factor graph faster leading to smooth estimation, it does not allow for taking advantage of the full FoV of the radar. The reduced FoV utilization when coupled with the proposed forward velocity only factor leads to poor velocity estimation in lateral and vertical directions. This is demonstrated when examining the body-frame linear velocity estimates from the radar-inertial only estimates~\cref{fig:v_all:gps_ri}. It can be seen that the forward velocity estimation tracks with the ground truth, but lateral and vertical velocity estimates drift as there is no information contributed from the radar measurements along those directions. One area of future research would be to incorporate radial speed measurements directly obtained from the radar into the factor graph. This has the advantages that information along lateral and vertical directions is added to the graph, as a function of how large the FoV is, as well as reducing approximations in the factor formulation.

\section{Conclusion}
This work presented ROAMER, a method which incorporates forward velocity information from a \ac{fmcw} radar to enhance the pose estimation robustness and velocity estimation accuracy during periods when the other sensing modalities, such as LiDAR odometry, are subject to performance degradation. The proposed method utilizes the direct velocity measurement capabilities of a radar sensor to augment the onboard LiDAR-inertial multimodal state estimation solution with measurements of forward velocity. This results in overall improved performance of the estimation solution despite periods of dropout of the LiDAR data. The proposed method was validated using hardware experiments conducted using an all-terrain vehicle driving at high speed in unstructured offroad environments.

\acknowledgments
The research was carried out at the Jet Propulsion Laboratory, California Institute of Technology, under a contract with the National Aeronautics and Space Administration (80NM0018D0004). This work was partially supported by Defense Advanced Research Projects Agency (DARPA).

\bibliographystyle{IEEEtran}
\bibliography{bib/general,bib/multi-modal,bib/radar,bib/lidar}

\thebiography

\begin{biographywithpic}
{Morten Nissov}{figures/biographies/mortenn_orig}
earned his B.S. in Electrical Engineering from the Technical University of Denmark in 2019 and M.S. in Electrical Engineering from the Technical University of Denmark in 2021. Currently, Morten is pursuing his Ph.D. in Engineering Cybernetics from the Norwegian University of Science and Technology. His current research is related to robot perception and estimation with focus on development of algorithms exploiting multi-sensor information in degraded conditions.
\end{biographywithpic}
\begin{biographywithpic}
{Shehryar Khattak}{figures/biographies/shehryar} is a Robotics Technologist within the Perception Systems Group at the NASA Jet Propulsion Laboratory. His work focuses on enabling resilient robot autonomy in complex environments through multi-sensor information fusion. Before joining JPL, he was a post-doctoral researcher at ETH Zurich and received his Ph.D. (2019) and MS
(2017) in Computer Science from the University of Nevada, Reno. He also holds an MS in Aerospace Engineering from KAIST (2012) and a BS in Mechanical Engineering from GIKI (2009). \end{biographywithpic}
\begin{biographywithpic}
{Jeffrey A. Edlund}{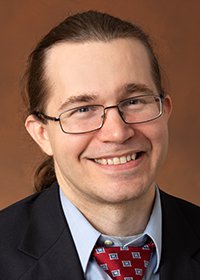}
received his B.S. degree in Physics in 2004 and a Ph.D. in Computation and Neural Systems from the California Institute of Technology in 2019. He is currently a Robotics Systems Engineer at the Jet Propulsion Laboratory, where he has worked on DARPA RACER, DARPA SUBT, Mars 2020, SAEL, and Axel.
\end{biographywithpic}
\begin{biographywithpic}
{Curtis Padgett}{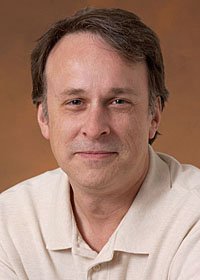}
is the Supervisor for the Perception Systems Group and a Principal in the Robotics Section at NASA’s Jet Propulsion Laboratory where he has worked on machine vision problems for over 30 years. He leads research efforts focused on aerial and maritime imaging problems including: navigation support for landing and proximity operations; path planning for sea surface vehicles using COLREGs; automated, real-time recovery of structure from motion; precision geo-registration of imagery; automated landmark generation and mapping for surface relative navigation; stereo image sea surface sensing for navigation on water and image based, multi-platform contact range determination. He has a Ph.D. in Computer Science from the University of California at San Diego. His research interests include pattern recognition, image-based reconstruction, and mapping.
\end{biographywithpic}
\begin{biographywithpic}
{Kostas Alexis}{figures/biographies/Kostas-Alexis-Photo-300x300}
is Full Professor at the Department of Engineering Cybernetics of the Norwegian University of Science and Technology (NTNU). Highlights of his research include leading Team CERBERUS winning the DARPA Subterranean Challenge and a host of contributions in the domain of resilient robotic autonomy – in perception, planning and control including learned navigation policies. Earlier research has included contributions in the ETH Zurich team that set the endurance world-record for UAVs in the below 50kg class with AtlantikSolar flying continuously for 81.5 hours. Dr. Alexis received his PhD in 2011 from the University of Patras in Greece. 
\end{biographywithpic}
\begin{biographywithpic}
{Patrick Spieler}{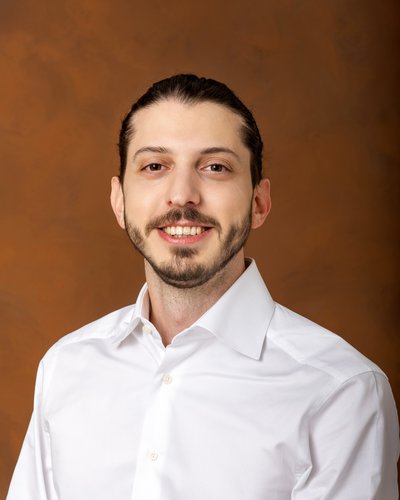}
is a Robotics Technologist within the Aerial Mobility Group at NASA Jet Propulsion Laboratory. He received his B.S. and M.S. degree in Robotics from Swiss Federal Institute of Technology, Lausanne (EPFL). Currently, he is the principal investigator of JPL’s team for the DARPA RACER project. Previously, he was a research engineer at California Institute of Technology (Caltech) where he led the Autonomous Flying Ambulance project and Leonardo, the first flying-walking robot. Before that he worked at iRobot and Astrocast, a space company building communication satellites.
\end{biographywithpic}

\end{document}